\documentclass[runningheads]{llncs}

\usepackage[T1]{fontenc}
\usepackage{amsmath,amssymb}
\usepackage{graphicx}
\usepackage{subcaption}
\usepackage{booktabs}
\usepackage{multirow}
\usepackage[table,xcdraw]{xcolor}
\usepackage[absolute,overlay]{textpos}
\usepackage{soul}
\usepackage{microtype}
\usepackage{wrapfig}

\definecolor{eb}{rgb}{0.12,0.49,0.85}
\definecolor{mint}{RGB}{0, 184, 186}
\definecolor{pink}{RGB}{229, 83, 171}
\definecolor{legorange}{RGB}{255,140,0}
\definecolor{leggreen}{RGB}{34,139,34}
\definecolor{violet}{RGB}{138,43,226}
\definecolor{darkyellow}{RGB}{184,134,11}
\definecolor{citeblue}{RGB}{0,51,153}

\sethlcolor{yellow!20}

\providecommand{\textul}[1]{\ul{#1}}
\newcommand{\cnum}[1]{\raisebox{.5pt}{\textcircled{\raisebox{-.9pt}{\scriptsize #1}}}}
\newcommand{\up}{\textcolor{red}{$\uparrow$}}
\newcommand{\down}{\textcolor{blue}{$\downarrow$}}

\usepackage[pagebackref,breaklinks,colorlinks,citecolor=citeblue,linkcolor=citeblue,urlcolor=citeblue]{hyperref}
\usepackage{orcidlink}  

\begin{document}

\title{Camera-to-Quadruped Robot Pose Estimation\\from Sequential Images}

\titlerunning{Camera-to-Quadruped Robot Pose Estimation}

\author{Seobin Hwang \and
Ji-Seong Beom \and
Yeong-Jun Cho\thanks{Corresponding author.}}

\authorrunning{S. Hwang et al.}

\institute{Department of Artificial Intelligence Convergence,\\
Chonnam National University, Gwangju, Republic of Korea\\
\email{\{cnu.cvl.hsb, haa0901\}@gmail.com, \{yj.cho\}@jnu.ac.kr}}

\maketitle

\begin{figure*}[h]
    \centering
    \includegraphics[width=0.96\linewidth]{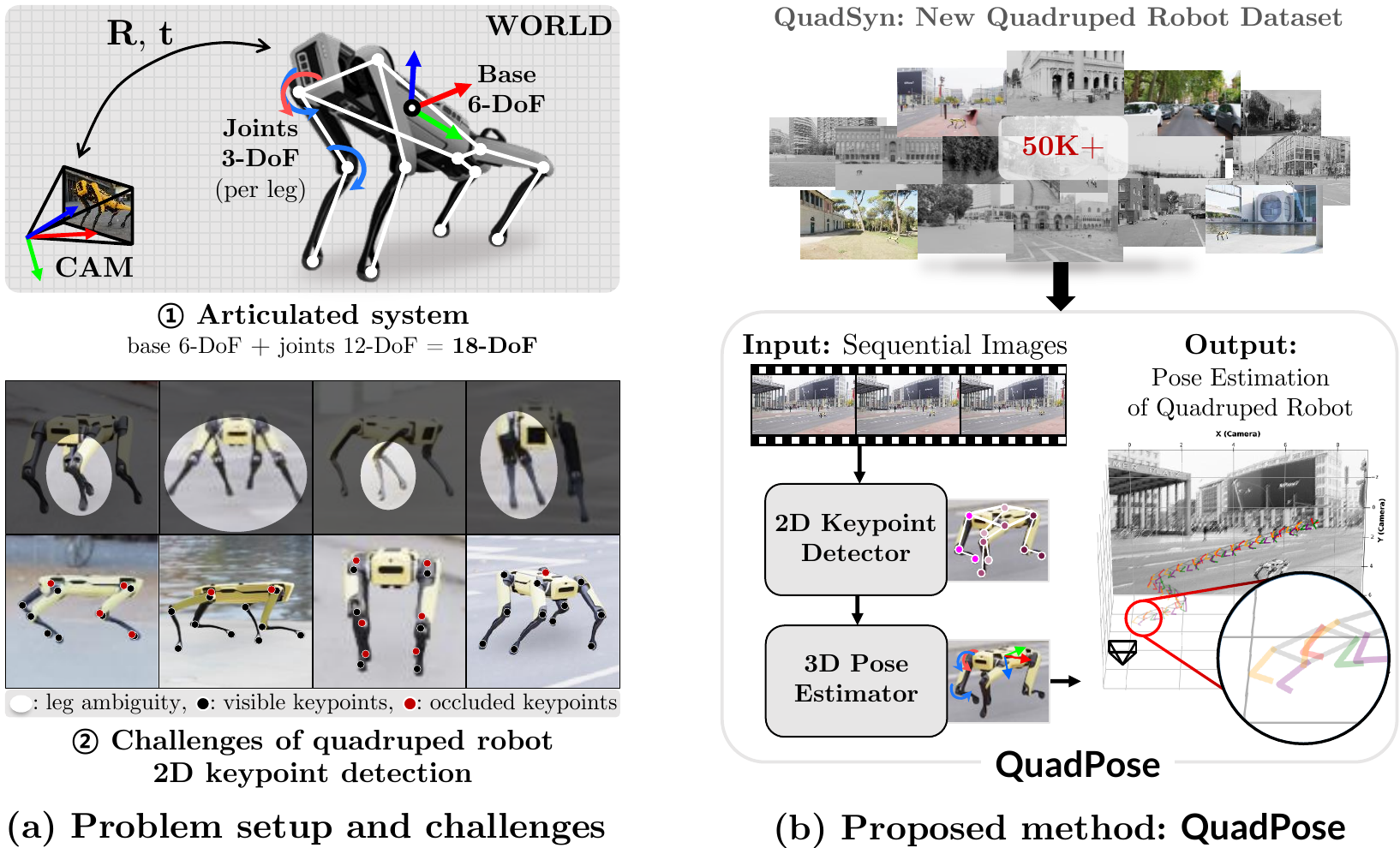}
    \caption{\textbf{Overview of camera-to-quadruped robot pose estimation.}
        (a) Problem setup and challenges: an articulated quadruped model (6-DoF base + 12-DoF joints) and 2D keypoint ambiguities caused by self-occlusion and leg symmetry.
        (b) Overview of the proposed \textsf{QuadPose} framework and the \texttt{QuadSyn} dataset.}
    \label{fig:01}
    \vspace{-30pt}
\end{figure*}

\begin{abstract}
Quadruped robots are increasingly deployed in hazardous environments such as disaster response and facility inspection, where external-camera monitoring is essential for safe operation.
This setting is challenging because quadrupeds are articulated systems with time-varying leg joints, leg symmetry, and frequent self-occlusions, which make keypoint correspondence ambiguous.
Moreover, no public benchmark exists, making quantitative comparison difficult.
To address these challenges, we propose \textsf{QuadPose}, which estimates the full-body 18-DoF pose (6-DoF base pose + 12-DoF joint angles) from monocular video sequences, and introduce \texttt{QuadSyn}, the first camera-to-quadruped robot pose benchmark.
On \texttt{QuadSyn}, \textsf{QuadPose} achieves 92.18\% PCK\text{@}$2.5$\,px, 95.54\% PCK-AUC, 85.94\% ADD@20\,cm, 66.98\% ADD-AUC, and 3.65$^\circ$ mean joint-angle error at approximately 33\,FPS.
\textit{Code and dataset will be publicly available.}
\end{abstract}

\section{Introduction}
\label{sec:introduction}

Quadruped robots are increasingly deployed for patrol, inspection, and monitoring in hazardous environments~\cite{fankhauser2018robust}, where real-time external-camera monitoring is essential for safe operation.
However, 2D tracking only provides the robot's location in the image plane, which is insufficient to assess whether it is stable, moving as intended, or approaching dangerous states.
Effective monitoring therefore requires estimating the robot's full-body 3D pose, including the 6-DoF base pose (position and orientation) and the articulated joint configuration.

\textls[-10]{
In practice, such monitoring often relies on fixed external cameras due to cost, maintenance, and reliability constraints.
As a result, only monocular video is available, without robot-side sensors (e.g., an IMU) or robot--camera extrinsic calibration.
Under these conditions, depth and 3D orientation remain inherently ambiguous from 2D boxes or keypoints, making it difficult to consistently relate image observations to the robot's kinematic state over time.
Estimating the robot's full-body 3D pose from a monocular camera remains a key challenge.
}

\textls[-20]{
This problem can be formulated as camera-to-robot pose estimation from external videos.
Existing approaches broadly fall into two categories: mesh-based and keypoint-based.
Mesh-based methods estimate pose by aligning the input image with rendered views of a CAD/mesh model~\cite{labbe2021single}.
Keypoint-based methods detect 2D keypoints and recover the 6-DoF base pose by solving a Perspective-n-Point (PnP) problem with known 3D keypoints~\cite{gao2003complete}, or by learning a direct mapping from images to pose~\cite{lee2020camera,lu2023markerless,tian2023robot,goswami2025robopepp}.
Most existing approaches target manipulators or drones, benefiting from fixed-base setups or rigid-body assumptions that simplify pose estimation.
}

\textls[-20]{
In contrast, as shown in Fig.~\ref{fig:01}(a), a quadruped has a 6-DoF base pose and time-varying leg joints, so a single 6-DoF pose is insufficient.
We estimate an 18-DoF state (6-DoF base + four 3-DoF legs), requiring 12 keypoints per frame (Fig.~\ref{fig:01}(a)--\cnum{1}).
Second, left--right and front--rear symmetry introduces ambiguity in keypoint correspondence, and self-occlusion during gait reduces the visual cues (Fig.~\ref{fig:01}(a)--\cnum{2}).
In addition, no public benchmark exists for camera-to-quadruped pose estimation, making fair comparison difficult.
Much of the prior work focuses on the robot's onboard sensing, while external-camera 3D pose estimation remains less explored.
}

To address this challenge, we propose \textsf{QuadPose}, which estimates the 18-DoF full-body pose of a quadruped robot from monocular video sequences (Fig.~\ref{fig:01}(b)).
Our key idea is to improve 2D keypoint quality, as accurate keypoints are critical for reliable 3D pose recovery.
Specifically, a keypoint-focused token selector mitigates leg appearance ambiguity caused by symmetric limbs.
In addition, skeleton-aware masked attention exploits kinematic connectivity to improve robustness under self-occlusion during gait (Fig.~\ref{fig:01}(a)--\cnum{2}).
Given stable 2D keypoints, \textsf{QuadPose} first predicts the 12-DoF joint angles and estimates the 6-DoF base pose, reconstructs 3D keypoints via forward kinematics, and then corrects residual errors using a refinement head.
Additionally, we introduce a new dataset \texttt{QuadSyn} to enable reproducible evaluation.

We evaluate \textsf{QuadPose} on the proposed \texttt{QuadSyn} benchmark and demonstrate consistent improvements over existing baselines.
Experimental results show that our keypoint-focused modeling improves 3D pose accuracy under symmetry and self-occlusion, while the refinement stage further reduces reconstruction errors.
\textsf{QuadPose} achieves 95.54\% PCK-AUC and 66.98\% ADD-AUC while running at about 33\,FPS.
In summary, our main contributions are as follows:
\begin{itemize}
\renewcommand{\labelitemi}{$\bullet$}
    \item \textls[-20]{\textbf{Full-Body Pose Estimation for Quadruped Robots.}}
        To our knowledge, we present the first camera-to-quadruped full-body pose estimation framework that recovers an 18-DoF state, including the 6-DoF base pose and 12 joints.
    \item \textls[-20]{\textbf{Robust 2D-to-3D Pose Recovery.}}
        We propose selector tokens and skeleton-aware masked attention to mitigate 2D keypoint ambiguity, while a refinement head further reduces 3D reconstruction errors.
    \item \textls[-20]{\textbf{New Quadruped Pose Dataset.}}
        To our knowledge, we build and release the first public camera-to-quadruped pose benchmark dataset.
\end{itemize}

\section{Related Works}
\label{sec:relatedworks}

\begin{table}[b!]
\setlength{\tabcolsep}{2pt}
\centering
\caption{\textbf{Overview of camera-to-robot pose estimation methods.}
Existing methods are categorized into mesh-based and keypoint-based approaches, and we summarize their target robot types, input modalities, and outputs.}
\label{tab:01}
\scriptsize
\begin{tabular}{clccc}
\hline
\textbf{Category} & \textbf{Method}    & \textbf{Robot type} & \textbf{Input} & \textbf{Output}* \\
\hline
 & RoboPose \textcolor{gray}{\tiny (CVPR'21)}~\cite{labbe2021single}      & Manipulator   & Image          & 6DoF + $q(7)$ \\
\multirow{-2}{*}{Mesh}
 & DronePose \textcolor{gray}{\tiny (ECCVW'20)}~\cite{albanis2020dronepose}& Drone         & Image          & 6DoF \\
\hline
 & DroneKey++ \textcolor{gray}{\tiny (ICRA'26)}~\cite{dronekey++} & Drone   & Image seq.          & 2D kpt + 6DoF \\
 & RoboPEPP \textcolor{gray}{\tiny (CVPR'25)}~\cite{goswami2025robopepp}  & Manipulator   & Image          & 2D/3D kpt + 6DoF + $q(7)$ \\
 & DroneKey \textcolor{gray}{\tiny (IROS'25)}~\cite{11247556}             & Drone         & Image seq. & 2D kpt + 6DoF \\
 & SGTAPose \textcolor{gray}{\tiny (CVPR'23)}~\cite{tian2023robot}        & Manipulator   & Image seq. & 2D/3D kpt + 6DoF \\
 & DREAM \textcolor{gray}{\tiny (ICRA'20)}~\cite{lee2020camera}           & Manipulator   & Image          & 2D/3D kpt + 6DoF \\
\multirow{-6}{*}{Keypoints}
 & \cellcolor[HTML]{EFEFEF}\textbf{\textsf{QuadPose} (Ours)}
 & \cellcolor[HTML]{EFEFEF}\textbf{Quadruped}
 & \cellcolor[HTML]{EFEFEF}\textbf{Image seq.}
 & \cellcolor[HTML]{EFEFEF}\textbf{2D/3D kpt + 6DoF + $q(12)$} \\ \hline
\end{tabular}
\\[0.05cm]
* 2D/3D kpt: 2D/3D keypoint coordinates, 6DoF: $R(3)+t(3)$ base pose, $q(J)$: joint angles.
\end{table}

\noindent \textbf{Camera-to-Robot Pose Estimation.} \;
Pose estimation aims to recover an object's position and orientation from visual observations and has been widely studied for humans and robots~\cite{lee2020camera,albanis2020dronepose,martinez2017simple}.
In 3D human pose estimation, body size varies across people, and monocular RGB suffers from scale ambiguity, so many methods predict root-relative or scale-normalized 3D pose~\cite{martinez2017simple,pavllo20193d}.
In contrast, robots have fixed dimensions and a known kinematic structure, making camera-to-robot pose estimation closer to recovering the state of a specific articulated robot in the camera frame.

\textls[-20]{
Camera-to-robot pose estimation recovers a robot's position and orientation in the camera frame from monocular RGB images~\cite{lee2020camera}.
For rigid platforms (e.g., drones and ground vehicles), the goal is to estimate a 6-DoF base transform, whereas articulated robots additionally require joint configurations.
For example, a quadruped requires estimating the 6-DoF base pose and the joint angles of its 12 leg joints.
Prior work can be grouped into mesh-based and keypoint-based methods, depending on whether they rely on 3D mesh models or keypoints as geometric priors (Table~\ref{tab:01}).
}

\noindent $\bullet$ \textbf{\textit{Mesh-based methods}}:
\textls[-10]{
Mesh-based approaches estimate pose by aligning the observed image with renderings of a robot CAD (3D mesh) model.
RoboPose~\cite{labbe2021single} iteratively compares rendered views with the input image to optimize the camera-to-robot pose and joint angles, while DronePose~\cite{albanis2020dronepose} leverages a 3D mesh prior for 6-DoF drone pose estimation.
These methods leverage strong geometric constraints but require iterative optimization at test time, resulting in slow inference.
}

\noindent $\bullet$ \textbf{\textit{Keypoint-based methods}}:
\textls[-10]{
Keypoint-based approaches detect 2D keypoints on robot joints, infer 3D joint locations using a kinematic model, and estimate pose from 2D--3D correspondences.
DREAM~\cite{lee2020camera} learns 2D keypoints from synthetic data with domain randomization~\cite{tobin2017domain} and estimates pose via PnP~\cite{gao2003complete}.
RoboPEPP~\cite{goswami2025robopepp} leverages pretraining to encode robot structure and improve robustness to occlusion.
Sequence-based methods also exploit temporal cues: DroneKey~\cite{11247556} and DroneKey++~\cite{dronekey++} target drones, while SGTAPose~\cite{tian2023robot} focuses on manipulators.
However, methods that estimate joint angles typically target fixed-base manipulators, whereas drones are treated as rigid bodies with no joints to estimate. To our knowledge, no prior work jointly estimates base motion and joint articulation.
}

\noindent \textbf{Datasets for Camera-to-Robot Pose Estimation.}\;
Existing datasets and evaluation protocols have mainly targeted manipulators and drones.
Manipulators benefit from fixed-base setups that simplify annotation~\cite{lee2020camera}, while drones can be treated as rigid bodies with straightforward 6-DoF labels~\cite{albanis2020dronepose,11247556,dronekey++}.
These datasets rely on synthetic data with background compositing for scalable ground-truth generation~\cite{lee2020camera,albanis2020dronepose,11247556,dronekey++}.
In contrast, to our knowledge, no public dataset or evaluation protocol exists for external-camera full-body pose estimation of quadruped robots, whose articulated structure and frequent self-occlusion make annotation substantially more challenging.
We address this gap in Sec.~\ref{sec:datasets}.

\section{Proposed Methods}
\label{sec:proposedmethods}

\begin{figure}[t!]
    \centering
    \includegraphics[width=1\linewidth]{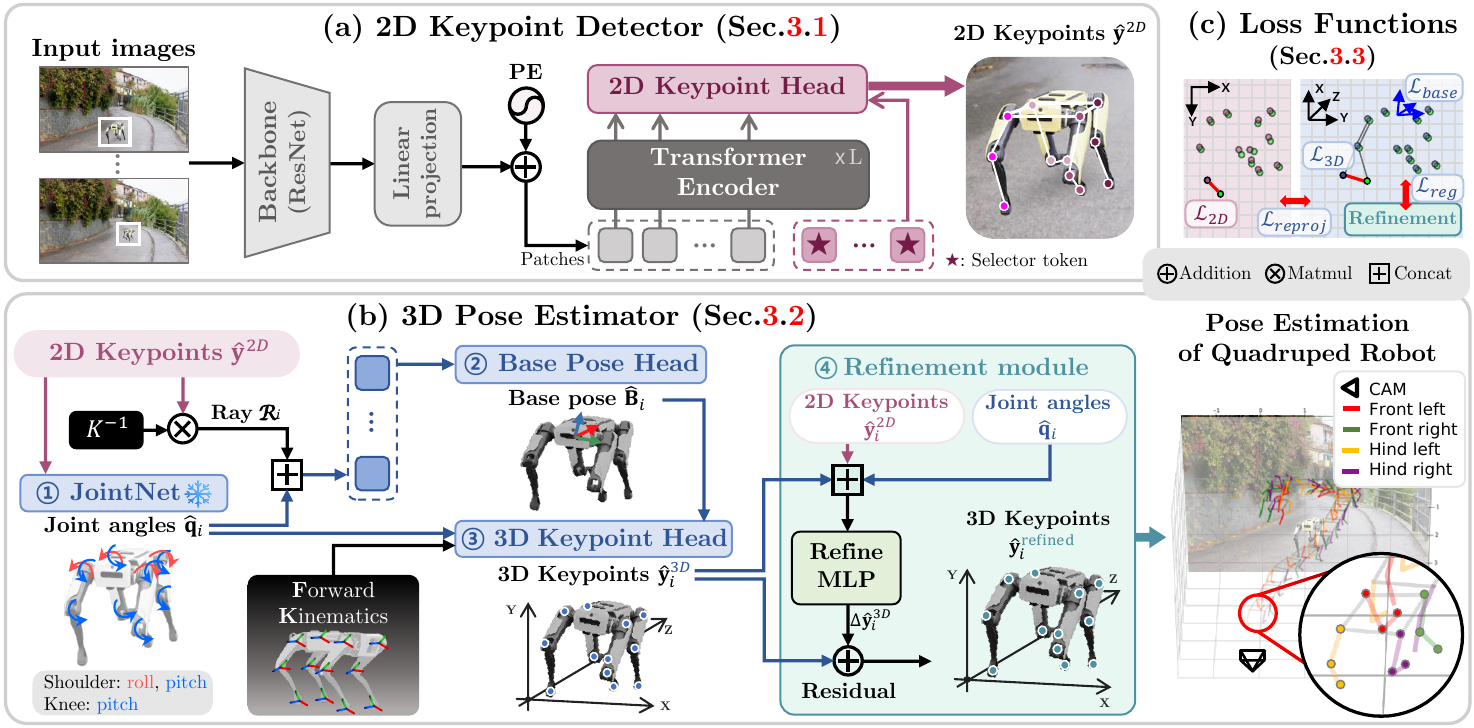}
    \caption{\textbf{The overall framework of \textsf{QuadPose}.}
        (a) 2D keypoint detector with selector tokens and skeleton-aware masked attention.
        (b) 3D pose estimator that predicts joint angles and base pose, with FK reconstruction and refinement.
        (c) Loss functions.}
    \label{fig:02}
\end{figure}

\textsf{QuadPose} consists of two main components: a 2D keypoint detector and a 3D pose estimator, as illustrated in Fig.~\ref{fig:02}.
Given a cropped robot bounding box, keypoint features are first extracted using a transformer-based~\cite{transformer} encoder.
These features are refined by a 2D keypoint head that incorporates cross-attention with selector tokens and a skeleton-aware masked attention module, enabling robust 2D keypoint detection.
The 3D pose estimator then predicts joint angles and the 6-DoF base pose from the detected 2D keypoints, reconstructs 3D keypoints via forward kinematics, and corrects residual errors through a refinement head.

Given an RGB image sequence $\{\mathbf{I}_i\}$ and camera intrinsics $\mathbf{K}$, the goal is to estimate the full-body 3D state of a quadruped robot in the camera frame at each time step $i$.
The robot is modeled as an articulated system with a rigid base and four multi-DoF legs comprising 12 joints, as shown in Fig.~\ref{fig:01}(a)--\cnum{1}.
The state comprises 18 degrees of freedom: the 6-DoF base pose $\mathbf{B}_i = (\mathbf{R}_i \in \mathbb{R}^{3 \times 3},\; \mathbf{t}_i \in \mathbb{R}^3)$ and the 12-DoF leg joint angles $\mathbf{q}_i \in \mathbb{R}^{12}$.
We assume that the URDF-based forward kinematics model~\cite{spot_urdf} and per-frame robot 2D bounding boxes are given.

\subsection{2D Keypoint Detector}
\label{sec:method:keypoint_detector}

The 2D keypoint detector estimates 12 keypoints corresponding to the quadruped robot's joints from an $H\times W$ input RGB image frame $\mathbf{I}_i\in \mathbb{R}^{H\times W \times 3}$.
It consists of a Transformer-based encoder~\cite{transformer} and a keypoint head as shown in Fig.~\ref{fig:02}(a).
The encoder extracts global contextual features, while the keypoint head predicts 2D keypoint coordinates.
To obtain a fixed-size input for the network, the robot bounding box is cropped and resized to a fixed resolution.

\noindent \textbf{Encoder.} \;
Global contextual information improves robustness to occlusion and background variations and enables effective feature sharing across the four structurally similar legs.
To capture such global spatial context over the entire image, we employ a ResNet backbone and a Transformer-based encoder.
\textls[-20]{
We first extract a feature map from the cropped robot image using a ResNet backbone~\cite{he2016deep}.
The resulting feature map is then linearly projected into a set of $N$ visual tokens of dimension $d$, yielding $\mathbf{F} \in \mathbb{R}^{N \times d}$.
Positional encodings $\mathbf{PE} \in \mathbb{R}^{N \times d}$ are added to the token features, which are then fed into the Transformer encoder:
}
\begin{equation}
    \mathbf{F}' = \text{Encoder}(\mathbf{F} + \mathbf{PE}).
\end{equation}
Here, $\mathbf{F}'$ represents context-aware visual token features that encode global spatial relationships across the image.

\begin{figure}[t]
    \centering
    \begin{minipage}[t]{0.45\linewidth}
        \centering
        \includegraphics[width=1\linewidth]{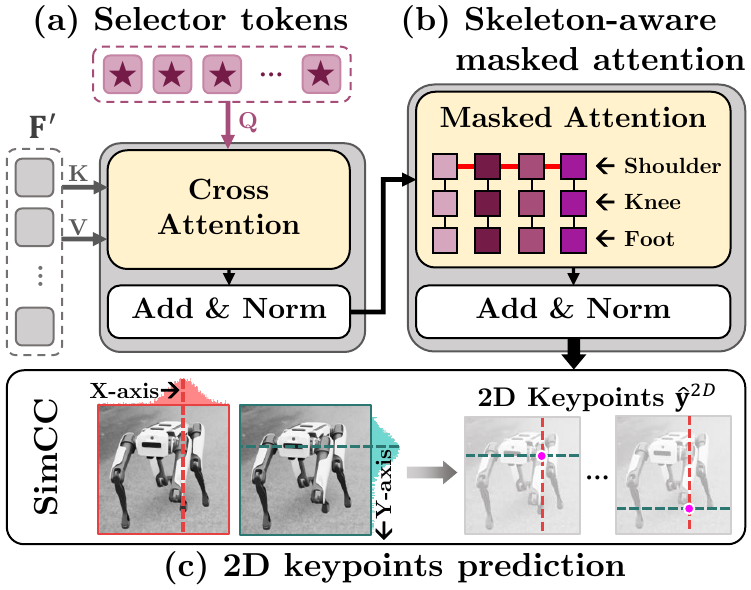}
        \caption{\textls[-20]{\textbf{Details of the 2D keypoint head.}
            It takes encoded visual tokens $\mathbf{F}'$ and predicts 2D keypoints via (a) selector-token cross-attention, (b) skeleton-aware masked self-attention, and (c) SimCC-based~\cite{simcc} 1D heatmap regression.}}
        \label{fig:03}
    \end{minipage}
    \hfill
    \begin{minipage}[t]{0.49\linewidth}
        \centering
        \includegraphics[width=1\linewidth]{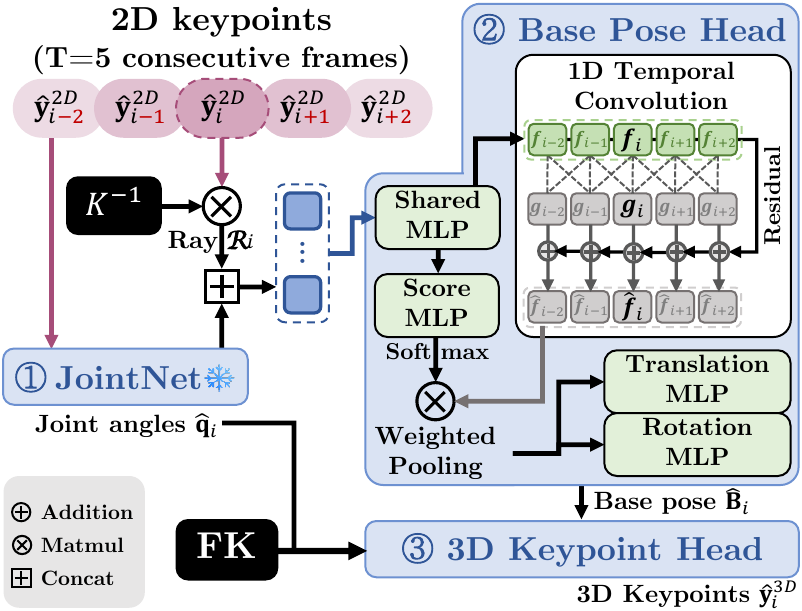}
        \caption{\textls[-15]{\textbf{Details of the 3D pose estimator.}
             \cnum{1} Given rays $\mathcal{R}_i$ and predicted joint angles $\hat{\mathbf{q}}_i$ over $T$ frames, it estimates the \cnum{2} 6-DoF base pose $\hat{\mathbf{B}}_i$ via temporal convolution and weighted pooling.
             \cnum{3} FK yields 3D keypoints $\hat{\mathbf{y}}_i^{3D}$ from $\hat{\mathbf{B}}_i$ and $\hat{\mathbf{q}}_i$.}}
        \label{fig:04}
    \end{minipage}
\end{figure}

\noindent \textbf{2D Keypoint Head.} \;
The 2D keypoint head localizes each of the 12 robot joints from the encoded visual features, as shown in Fig.~\ref{fig:03}.

\noindent $\bullet$ \textbf{\textit{Selector tokens}:}
Although the four legs of a quadruped robot share a similar kinematic structure, each joint occupies a distinct spatial location and plays a different functional role.
Relying solely on shared global features $\mathbf{F}'$ may therefore lead to ambiguity when localizing individual keypoints.
To address this issue, we introduce a set of learnable selector tokens that explicitly guide the network to attend to joint-specific regions.

Specifically, we define $12$ selector tokens $\mathbf{S} = \{\mathbf{s}_1, \ldots, \mathbf{s}_{12}\} \in \mathbb{R}^{12 \times d}$, each corresponding to a distinct joint of the quadruped robot.
These selector tokens interact with the encoded visual tokens $\mathbf{F}'$ via a cross-attention mechanism (Fig.~\ref{fig:03}(a)), where $\mathbf{S}$ serves as the queries and the encoded feature tokens $\mathbf{F}'$ act as the keys and values as follows:
\begin{equation}
    \mathbf{Z} = \text{CrossAttn}(Q=\mathbf{S}, K=\mathbf{F}', V=\mathbf{F}'),
\end{equation}
\textls[-10]{
The output $\mathbf{Z} \in \mathbb{R}^{12 \times d}$ aggregates joint-specific information, enabling each selector token to focus on its corresponding keypoint while leveraging global context.
}

\noindent $\bullet$ \textbf{\textit{Skeleton-aware masked attention}:}
\textls[-25]{
While the proposed selector tokens encourage joint-specific attention, challenges remain due to the highly similar appearance of the legs and frequent self-occlusion, as illustrated in Fig.~\ref{fig:01}(a)--\cnum{2}.
To better handle these, we leverage the quadruped robot's kinematic topology as a structural prior.
This topology is represented by a skeleton adjacency matrix $\mathbf{A} \in \{0,1\}^{12 \times 12}$ over the 12 keypoints, corresponding to the shoulder--knee--foot kinematic chain of each leg.
}
We use $\mathbf{A}$ as an attention mask and apply masked self-attention to $\mathbf{Z}$:
\begin{equation}
    \mathbf{Z}' = \text{MaskedSelfAttn}(\mathbf{Z}, \; \; \text{mask}=\mathbf{A}) \in \mathbb{R}^{12 \times d}.
\end{equation}
Here, information exchange is restricted to directly adjacent joints along the kinematic chain (e.g., front-left shoulder $\leftrightarrow$ front-left knee $\leftrightarrow$ front-left foot), as illustrated in Fig.~\ref{fig:03}(b). Non-adjacent joints (e.g., front-left knee and front-right foot) do not directly exchange information.
This skeleton-aware masked self-attention enables topology-consistent information propagation, improving robustness to leg ambiguity and joint self-occlusion.

\noindent $\bullet$ \textbf{\textit{2D keypoints prediction}:}
Many existing methods~\cite{toshev2014deeppose,mao2022poseur} perform 2D keypoint prediction via direct coordinate regression; however, this approach is sensitive to noise and does not model prediction uncertainty.
Instead of direct regression, we predict independent 1D distributions along the $x$ and $y$ axes to jointly estimate keypoint coordinates and confidence, as shown in Fig.~\ref{fig:03}(c).
To this end, we adopt the SimCC formulation~\cite{simcc} to implement this design.
Specifically, for each $d$-dimensional keypoint token $\mathbf{z}'_k \in \mathbb{R}^d$ (the $k$-th token in $\mathbf{Z}'$), we predict independent 1D probability distributions along the $x$- and $y$-axes as follows:
\begin{equation}
    p_k^x = \text{Softmax}\left(\text{MLP}_x(\mathbf{z}_k')\right) \in \mathbb{R}^W, \quad
    p_k^y = \text{Softmax}\left(\text{MLP}_y(\mathbf{z}_k')\right) \in \mathbb{R}^H,
\end{equation}
\textls[-20]{
where $\text{MLP}_x$ and $\text{MLP}_y$ denote multi-layer perceptrons used to predict the 1D distributions along the $x$ and $y$ axes, respectively.
We then compute the $(x,y)$ 2D coordinates as the expectation of each 1D distribution:
$\hat{x}_k = \sum_{u=1}^{W} u \cdot p_k^x(u),$ and $\hat{y}_k = \sum_{v=1}^{H} v \cdot p_k^y(v)$.
Finally, the detector outputs the 2D keypoints $\hat{\mathbf{y}}^{2D} = \{(\hat{x}_k, \hat{y}_k)\}_{k=1}^{12}$.
}

\subsection{3D Pose Estimator}
\label{sec:method:pose_estimator}

The 3D pose estimator infers the quadruped's 3D pose in the camera coordinate frame from the detected 2D keypoints $\hat{\mathbf{y}}^{2D}$ and the camera intrinsics matrix $\mathbf{K}$.
The estimator consists of four components: \cnum{1} JointNet, \cnum{2} Base Pose Head, \cnum{3} 3D Keypoint Head, and \cnum{4} Refinement module, as illustrated in Fig.~\ref{fig:02}(b).

\noindent \textbf{JointNet.} \;
A lightweight MLP (JointNet) predicts the 12-DoF leg joint angles $\hat{\mathbf{q}}_i \in \mathbb{R}^{12}$ from the detected 2D keypoints $\hat{\mathbf{y}}^{2D}_i$.
This module is pretrained independently and kept \emph{frozen} during end-to-end training of the full pipeline.
This JointNet design prevents early-stage errors from propagating to downstream modules and simplifies the learning problem.
With JointNet kept frozen, the forward kinematics (FK) chain becomes a deterministic function of the base pose given the fixed joint angles $\hat{\mathbf{q}}_i$. Specifically, FK computes the 3D joint positions by chaining rigid transformations along the robot's kinematic chain (e.g., shoulder$\rightarrow$knee$\rightarrow$foot) given the joint angles and base pose.

\noindent \textbf{Base Pose Head.} \;
The base pose head estimates the 6-DoF base pose $\hat{\mathbf{B}}_i = (\hat{\mathbf{R}}_i, \hat{\mathbf{t}}_i)$ from the detected 2D keypoints $\hat{\mathbf{y}}^{2D}_i = \{(\hat{x}_k, \hat{y}_k)\}_{k=1}^{12}$ and the predicted joint angles $\hat{\mathbf{q}}_i$.
The overall structure of the base pose head is shown in Fig.~\ref{fig:04}.
To convert the detected keypoints to camera coordinates, we first remap $\hat{\mathbf{y}}^{2D}_i$ to the original image coordinates and normalize them using the inverse intrinsic matrix:
\begin{equation}
    \mathbf{r}_k = \mathbf{K}^{-1}
    [\hat{x}_k, \hat{y}_k, 1]^\top \in \mathbb{R}^3,
\end{equation}
where $\mathbf{r}_k$ denotes the 3D ray from the camera center toward the $k$-th keypoint.
At frame $i$, we denote the set of ray vectors as $\mathcal{R}_i=\left[\mathbf{r}_{1};\ldots;\mathbf{r}_{12}\right]\in \mathbb{R}^{12\times 3}$.
We then concatenate $\mathcal{R}_i$ with the predicted joint angles $\hat{\mathbf{q}}_i\in \mathbb{R}^{12}$ to form the per-frame input $\left[\mathcal{R}_i, \;\hat{\mathbf{q}}_i\right]\in \mathbb{R}^{12\times 4}$ to the base pose head.
A single frame provides only a 2D projection and thus cannot determine the robot's absolute distance (scale) from the camera.
In contrast, temporal cues across frames, such as consistent scale trends, provide additional evidence about depth changes.
We therefore incorporate $T$ consecutive frames to help disambiguate this depth-scale.

However, not all frames are equally informative. When the robot moves, changes in its apparent scale and viewpoint provide stronger depth and translation cues, whereas near-stationary frames tend to be less informative for resolving 3D ambiguity.
To account for this, we employ learnable weighted pooling for temporal aggregation.
At each frame $i$, $\mathcal{R}_i$ and $\hat{\mathbf{q}}_i$ are concatenated and independently passed through a shared MLP, producing $T$ per-frame features $\{\mathbf{f}_i\}_{i=1}^{T}$.
These features are fed into 1D temporal convolutions to capture how features change across neighboring frames~\cite{pavllo20193d}, yielding $\mathbf{g}_i$.
A residual connection yields $\hat{\mathbf{f}}_i = \mathbf{g}_i + \mathbf{f}_i$, preserving per-frame cues that may be smoothed out by the convolution.
\textls[-10]{
To prioritize informative frames, a Score MLP predicts a scalar score for each $\hat{\mathbf{f}}_i$, normalized via softmax to obtain per-frame weights.
Weighted pooling produces a fused feature, which two MLP heads decode into the base translation $\hat{\mathbf{t}}_i \in \mathbb{R}^3$ and a 6D rotation representation~\cite{zhou2019continuity}, converted to a $3 \times 3$ rotation matrix $\hat{\mathbf{R}}_i$.
}

\noindent \textbf{3D Keypoint Head.} \;
Given the predicted base pose $\hat{\mathbf{B}}_i=(\hat{\mathbf{R}}_i,\hat{\mathbf{t}}_i)$ and joint angles $\hat{\mathbf{q}}_i$, we compute the 3D keypoints $\hat{\mathbf{y}}_i^{3D} \in \mathbb{R}^{12 \times 3}$ via URDF-based forward kinematics (FK)~\cite{spot_urdf}.
FK applies rigid transformations sequentially along each kinematic chain:
starting from the base frame, it computes the shoulder, knee, and foot positions for each leg using the URDF-defined kinematic parameters and the joint angles.
We refer to the FK output as the initial 3D keypoints $\hat{\mathbf{y}}_i^{3D}$.

\noindent \textbf{Refinement Module.} \;
While FK enforces kinematic consistency, errors from upstream modules can propagate to the initial 3D keypoints.
To correct these, we concatenate the coarse 3D keypoints $\hat{\mathbf{y}}_i^{3D}$, the 2D keypoints $\hat{\mathbf{y}}_i^{2D}$, and the predicted joint angles $\hat{\mathbf{q}}_i$ into a single input vector,
and pass it through a Refine MLP that predicts a per-keypoint residual offset $\Delta \hat{\mathbf{y}}_i^{3D}$.
The refined 3D keypoints are then obtained by $\hat{\mathbf{y}}_i^{\text{refined}} = \hat{\mathbf{y}}_i^{3D} + \Delta \hat{\mathbf{y}}_i^{3D}$.

\subsection{Loss Functions}
\label{sec:method:loss}

We train the 2D keypoint detector and 3D pose estimator independently.
Within the 3D pose estimator, JointNet is pretrained and frozen before training the remaining components.

\noindent \textbf{2D keypoint detector loss.} \;
The 2D keypoint detector follows the SimCC~\cite{simcc} formulation and uses KL divergence between the predicted 1D distributions and Gaussian-smoothed GT distributions along the $x$ and $y$ axes as the loss $\mathcal{L}_{\text{2D}}$.

\noindent \textbf{3D pose estimator loss.} \;
The 3D pose estimator uses three loss components.

\noindent $\bullet$ \textbf{\textit{Kinematic loss:}}
\textls[-15]{
$\mathcal{L}_{\text{joint}}$ is the $\ell_1$ loss on the predicted joint angles $\hat{\mathbf{q}}_i$, and
$\mathcal{L}_{\text{3D}}$ is the $\ell_1$ loss on the coarse 3D keypoints $\hat{\mathbf{y}}_i^{3D}$, both derived from the FK chain:
}
\begin{equation}
    \mathcal{L}_{\text{kin}} = \lambda_{\text{joint}}\mathcal{L}_{\text{joint}} + \lambda_{\text{3D}}\mathcal{L}_{\text{3D}}.
\end{equation}

\noindent $\bullet$ \textbf{\textit{Base pose loss}:}
$\mathcal{L}_{\text{trans}}$ is the $\ell_1$ loss on base translation $\hat{\mathbf{t}}_i$, and $\mathcal{L}_{\text{rot}}$ is the geodesic loss~\cite{zhou2019continuity} on base rotation $\hat{\mathbf{R}}_i$:
\begin{equation}
    \mathcal{L}_{\text{base}} = \lambda_{\text{trans}}\mathcal{L}_{\text{trans}} + \lambda_{\text{rot}}\mathcal{L}_{\text{rot}}.
\end{equation}

\noindent $\bullet$ \textbf{\textit{Refinement loss}:}
\textls[-20]{
$\mathcal{L}_{\text{refined}}$ is the $\ell_1$ loss on the refined 3D keypoints $\hat{\mathbf{y}}_i^{\text{refined}}$.
$\mathcal{L}_{\text{reproj}}$ is the $\ell_2$ reprojection error between the projected refined 3D keypoints and the detected 2D keypoints.
$\mathcal{L}_{\text{reg}}$ is the $\ell_2$ penalty on the residual offset magnitude $\|\Delta \hat{\mathbf{y}}_i^{3D}\|$, preventing the refinement module from making unnecessarily large corrections:
}
\begin{equation}
    \mathcal{L}_{\text{refine}} = \lambda_{\text{refined}}\mathcal{L}_{\text{refined}} + \lambda_{\text{reproj}}\mathcal{L}_{\text{reproj}} + \lambda_{\text{reg}}\mathcal{L}_{\text{reg}}.
\end{equation}
The final 3D pose estimator loss is: $\mathcal{L}_{\text{3D-pose}} = \mathcal{L}_{\text{kin}} + \mathcal{L}_{\text{base}} + \mathcal{L}_{\text{refine}}.$

\section{\texttt{QuadSyn}: A New Benchmark for Quadruped Robots}
\label{sec:datasets}

\begin{figure}[b!]
    \centering
    \includegraphics[width=0.98\linewidth]{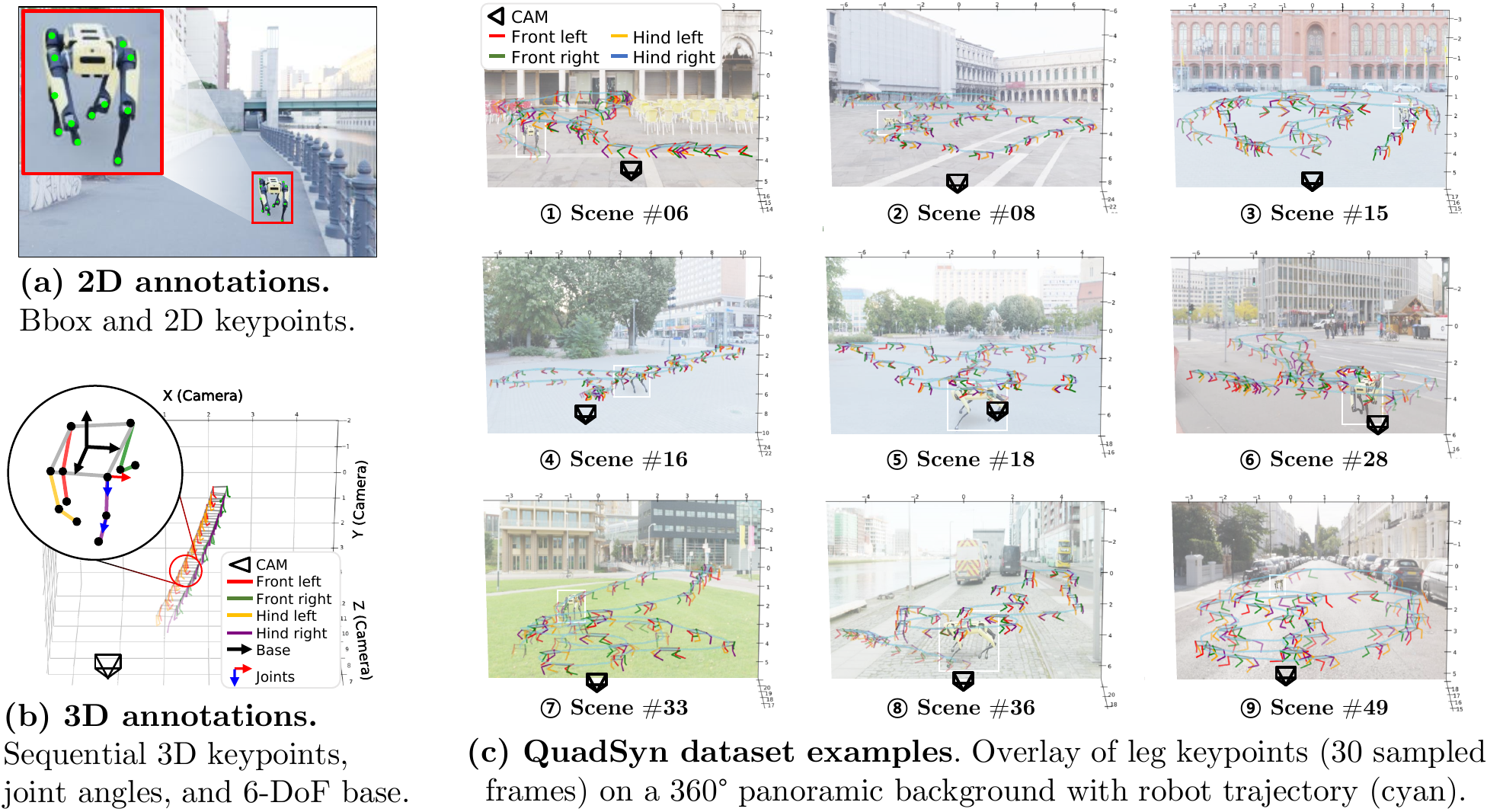}
    \caption{\textbf{\texttt{QuadSyn} dataset annotations and examples.}
    \textls[-20]{
    (a) 2D bounding box and keypoint labels.
    (b) 3D keypoints, joint angles, and 6-DoF base pose in the camera frame.
    (c) Scene examples with overlaid leg keypoint trajectories (30 frames) and robot paths.}}
    \label{fig:05}
\end{figure}

We introduce \texttt{QuadSyn}, to our knowledge, the first benchmark dataset for 18-DoF camera-to-quadruped robot pose estimation.
Quadruped robots are increasingly deployed in real-world scenarios such as disaster response and facility inspection, where images are captured under diverse and challenging conditions. However, there is no publicly available dataset that provides real camera images of quadruped robots together with the key labels needed for pose estimation (e.g., 2D/3D joint annotations and camera setup information).
Attaching markers alters the robot's appearance, and synchronizing 3D labels from external cameras during free locomotion remains difficult.
To fill this gap, we build \texttt{QuadSyn}. Although \texttt{QuadSyn} is synthetic, we design its data pipeline to make rendered images closely match real camera observations, thereby reducing the domain gap.

\textls[-10]{
\texttt{QuadSyn} contains 51,697 synthetic RGB images at a resolution of $1280\times720$, generated using a single quadruped robot (Boston Dynamics Spot) rendered over 52 background scenes.
Each scene forms a video sequence of about 1,000 frames recorded at 30 FPS, resulting in 52 sequences in total.
We generate \texttt{QuadSyn} using NVIDIA Isaac Sim~\cite{isaacsim}.
To mimic real-world visual contexts, we build 360$^\circ$ panoramic textures from collected real-world images and apply them as environment backgrounds in the simulator.
This reduces the synthetic-to-real domain gap; Section~\ref{sec:harper} confirms that our model transfers to real data without domain adaptation.
We further simulate walking motions with the physics engine and vary robot trajectories (including turns), gait patterns, and camera--robot relative poses across scenes to cover a wide range of motions and viewpoints.
Additional analysis of the domain gap and dataset is provided in the supplementary material.
}

For each frame, we derive the following annotations from the simulator-provided robot states (base pose and joint angles) and camera parameters (Fig.~\ref{fig:05}):
\begin{itemize}
\renewcommand{\labelitemi}{$\bullet$}
    \item 2D/3D keypoints for 12 leg joints (shoulder--knee--foot for all 4 legs),
    \item the base 6-DoF pose in the camera frame ($\mathbf{R}$, $\mathbf{t}$) and 12-DoF joint angles,
    \item camera intrinsics and extrinsics,
    \item the 2D bounding box of the robot in each frame.
\end{itemize}

Figure~\ref{fig:05}(a) shows the 2D bounding box and 2D keypoint annotations, and Fig.~\ref{fig:05}(b) visualizes the 3D keypoints along with the robot base pose (6-DoF) and joint angles in the camera and world frames.
In particular, we compute 2D keypoints by transforming 3D keypoints into the camera frame and projecting them using the camera projection model, ensuring 2D--3D correspondence.
Example scenes and 3D keypoint trajectories are shown in Fig.~\ref{fig:05}(c), and additional details of the data generation process are included in the supplementary video.

\section{Experimental Results}
\label{sec:experimentalresults}

\subsection{Settings}
\label{sec:experiment:settings}

\noindent \textbf{Datasets.} \;
We conduct our main quantitative experiments on \texttt{QuadSyn}, a synthetic benchmark dataset.
As described in Sec.~\ref{sec:datasets}, \texttt{QuadSyn} contains outdoor scenes with diverse backgrounds, motions, and pose variations.
It provides annotations for 2D/3D keypoints, the robot base pose (6-DoF), and joint angles.
For evaluation in real-world settings, we additionally use \texttt{Harper}~\cite{avogaro2024harper}, which is captured from a fixed external camera, to evaluate cross-domain generalization qualitatively.
\texttt{Harper} is an HRI dataset primarily collected from the robot's viewpoint for human pose analysis. It also provides videos captured from an external camera, where the quadruped robot is visible. However, the external-view videos do not include camera intrinsics or robot annotations. Therefore, we manually annotate 2D keypoints and restrict our evaluation to 2D keypoint accuracy.

\noindent \textbf{Implementation details.} \;
\textls[-20]{
All experiments are implemented using PyTorch~\cite{paszke2019pytorch} and trained on an A100 GPU using the Adam optimizer~\cite{kingma2014adam} for 100 epochs.
The full pipeline runs at $\sim$33\,FPS with 34.7M parameters.
The batch size and other hyperparameters are empirically determined to balance accuracy and computational cost.
For the pose estimation network, the loss weights are set as
$\lambda_{\text{joint}} = 0.1$,
$\lambda_{\text{3D}} = 50.0$,
$\lambda_{\text{trans}} = 10.0$,
$\lambda_{\text{rot}} = 1.0$,
$\lambda_{\text{refined}} = 100.0$,
$\lambda_{\text{reproj}} = 0.1$, and
$\lambda_{\text{reg}} = 0.5$.
}

\noindent \textbf{Evaluation metrics.} \;
We evaluate our method in three aspects: 2D keypoint localization, 3D base pose estimation, and joint angle prediction.
For 2D keypoint localization, we use the percentage of correct keypoints (PCK)~\cite{tremblay2018synthetically} with pixel thresholds of 2.5, 5, and 10.
We also report the area under the PCK curve (AUC) and the median pixel error.
Base pose accuracy is evaluated using the average distance metric (ADD)~\cite{lee2020camera}.
ADD computes the mean Euclidean distance between ground-truth 3D keypoints and those transformed by the predicted camera-to-robot pose.
Joint configuration accuracy is measured by the mean absolute error (MAE) in degrees, averaged over all samples and 12 joints.

For real-world evaluation on \texttt{Harper}~\cite{avogaro2024harper}, no ground-truth 2D keypoints are available for the external view; therefore, we manually annotate keypoints.
To account for potential annotation uncertainty, we adopt OKS-based mAP~\cite{lin2014microsoft}, which provides a soft, tolerance-aware accuracy measure.
We additionally measure temporal jitter, defined as frame-to-frame variations in keypoint positions, to assess temporal stability.

\subsection{Results on \texttt{QuadSyn} Dataset}
\label{sec:exp:resultonquadsyn}

\begin{table}[b!]
\scriptsize
\renewcommand{\arraystretch}{1.0}
\setlength{\tabcolsep}{1.5pt}
\centering
\caption{\textbf{PCK and ADD comparison on the \texttt{QuadSyn} benchmark.}
We compare with state-of-the-art methods~\cite{lee2020camera,labbe2021single,yolov8-pose,tian2023robot,goswami2025robopepp} and report PCK at 2.5/5/10\,px and ADD at 10/15/20\,cm, each with AUC and median error ($\uparrow$: higher is better, $\downarrow$: lower is better). Missing entries (--) indicate that the method does not produce the corresponding output.}
\label{tab:quadsyn_exp_pck-add}
\begin{tabular}{lcccccccccc}
\hline \noalign{\hrule height 0.5pt}
 & \multicolumn{5}{c}{\textbf{PCK @ (pix)}} & \multicolumn{5}{c}{\textbf{ADD @ (cm)}} \\
\multirow{-2}{*}{\textbf{Method}} & \textbf{2.5}\up & \textbf{5}\up & \textbf{10}\up & \textbf{AUC}\up & \textbf{Med}\down & \textbf{10}\up & \textbf{15}\up & \textbf{20}\up & \textbf{AUC}\up & \textbf{Med}\down \\ \hline \noalign{\hrule height 0.5pt}
DREAM \textcolor{gray}{\tiny (ICRA'20)}~\cite{lee2020camera} & 24.45 & 75.62 & 98.52 & 61.13 & 3.720 & 19.10 & 28.20 & 35.70 & 8.400 & 31.93 \\
RoboPose \textcolor{gray}{\tiny (CVPR'21)}~\cite{labbe2021single} & -- & -- & -- & -- & -- & 0.120 & 0.450 & 0.720 & 0.004 & 732.6 \\
YOLOv8-pose \textcolor{gray}{\tiny ('23)}~\cite{yolov8-pose} & 70.44 & 92.54 & 98.97 & \textul{88.90} & 1.683 & -- & -- & -- & -- & -- \\
SGTAPose \textcolor{gray}{\tiny (CVPR'23)}~\cite{tian2023robot} & \textul{86.58} & \textul{96.55} & \textul{99.15} & 87.04 & \textul{0.740} & \textul{52.08} & \textul{63.93} & 72.12 & \textul{56.97} & \textul{9.270} \\
RoboPEPP \textcolor{gray}{\tiny (CVPR'25)}~\cite{goswami2025robopepp} & 70.53 & 86.81 & 96.06 & 75.99 & 1.410 & 23.58 & 55.18 & \textul{85.87} & 51.63 & 14.03 \\
\cellcolor[HTML]{EFEFEF}\textbf{\textsf{QuadPose} (Ours)} & \cellcolor[HTML]{EFEFEF}\textbf{92.18} & \cellcolor[HTML]{EFEFEF}\textbf{97.57} & \cellcolor[HTML]{EFEFEF}\textbf{99.58} & \cellcolor[HTML]{EFEFEF}\textbf{95.54} & \cellcolor[HTML]{EFEFEF}\textbf{0.407} & \cellcolor[HTML]{EFEFEF}\textbf{69.49} & \cellcolor[HTML]{EFEFEF}\textbf{81.12} & \cellcolor[HTML]{EFEFEF}\textbf{85.94} & \cellcolor[HTML]{EFEFEF}\textbf{66.98} & \cellcolor[HTML]{EFEFEF}\textbf{6.740} \\
\cellcolor[HTML]{EFEFEF}\textit{Improvement*} & \cellcolor[HTML]{EFEFEF}\textit{\textcolor{red}{+5.60}} & \cellcolor[HTML]{EFEFEF}\textit{\textcolor{red}{+1.02}} & \cellcolor[HTML]{EFEFEF}\textit{\textcolor{red}{+0.43}} & \cellcolor[HTML]{EFEFEF}\textit{\textcolor{red}{+6.64}} & \cellcolor[HTML]{EFEFEF}\textit{\textcolor{blue}{-0.33}} & \cellcolor[HTML]{EFEFEF}\textit{\textcolor{red}{+17.4}} & \cellcolor[HTML]{EFEFEF}\textit{\textcolor{red}{+17.2}} & \cellcolor[HTML]{EFEFEF}\textit{\textcolor{red}{+0.07}} & \cellcolor[HTML]{EFEFEF}\textit{\textcolor{red}{+10.0}} & \cellcolor[HTML]{EFEFEF}\textit{\textcolor{blue}{-2.53}} \\ \hline \noalign{\hrule height 0.5pt}
\end{tabular}

* Improvement is computed against the best baseline (excluding ours).
\end{table}

\begin{table}[b!]
\scriptsize
\renewcommand{\arraystretch}{1.0}
\setlength{\tabcolsep}{2.5pt}
\centering
\caption{\textls[-20]{\textbf{Joint-angle error comparison on the \texttt{QuadSyn} benchmark.}
We compare with RoboPose~\cite{labbe2021single} and RoboPEPP~\cite{goswami2025robopepp} and report per-joint angle error ($\downarrow$: lower is better).}}
\label{tab:quadsyn_exp_joint}
\begin{tabular}{lcccccccccccc}
\hline \noalign{\hrule height 0.5pt}
 & \multicolumn{12}{c}{\textbf{Joint Error @ (degree)} \down} \\
 & \multicolumn{3}{c}{\textbf{Front-Left}} & \multicolumn{3}{c}{\textbf{Front-Right}} & \multicolumn{3}{c}{\textbf{Rear-Left}} & \multicolumn{3}{c}{\textbf{Rear-Right}} \\
\multirow{-3}{*}{\textbf{Method}} & \textbf{J1} & \textbf{J2} & \textbf{J3} & \textbf{J4} & \textbf{J5} & \textbf{J6} & \textbf{J7} & \textbf{J8} & \textbf{J9} & \textbf{J10} & \textbf{J11} & \textbf{J12} \\ \hline \noalign{\hrule height 0.5pt}
RoboPose \textcolor{gray}{\tiny (CVPR'21)}~\cite{labbe2021single} & 10.6 & 10.6 & 9.56 & 9.45 & 19.9 & 14.5 & 13.1 & 19.2 & 12.7 & 25.0 & 13.2 & 8.98 \\
RoboPEPP \textcolor{gray}{\tiny (CVPR'25)}~\cite{goswami2025robopepp} & \textul{4.11} & \textul{4.46} & \textbf{4.01} & \textul{3.85} & \textul{3.85} & \textul{3.99} & \textul{4.14} & \textul{3.61} & \textul{4.09} & \textul{4.40} & \textul{3.97} & \textul{4.33} \\
\cellcolor[HTML]{EFEFEF}\textbf{\textsf{QuadPose} (Ours)} & \cellcolor[HTML]{EFEFEF}\textbf{3.45} & \cellcolor[HTML]{EFEFEF}\textbf{4.07} & \cellcolor[HTML]{EFEFEF}\textul{4.30} & \cellcolor[HTML]{EFEFEF}\textbf{3.42} & \cellcolor[HTML]{EFEFEF}\textbf{3.22} & \cellcolor[HTML]{EFEFEF}\textbf{3.50} & \cellcolor[HTML]{EFEFEF}\textbf{3.54} & \cellcolor[HTML]{EFEFEF}\textbf{3.14} & \cellcolor[HTML]{EFEFEF}\textbf{3.83} & \cellcolor[HTML]{EFEFEF}\textbf{3.75} & \cellcolor[HTML]{EFEFEF}\textbf{3.59} & \cellcolor[HTML]{EFEFEF}\textbf{3.99} \\
\cellcolor[HTML]{EFEFEF}\textit{Improvement*} & \cellcolor[HTML]{EFEFEF}\textit{\textcolor{blue}{-0.7}} & \cellcolor[HTML]{EFEFEF}\textit{\textcolor{blue}{-0.4}} & \cellcolor[HTML]{EFEFEF}\textit{\textcolor{red}{+0.3}} & \cellcolor[HTML]{EFEFEF}\textit{\textcolor{blue}{-0.4}} & \cellcolor[HTML]{EFEFEF}\textit{\textcolor{blue}{-0.6}} & \cellcolor[HTML]{EFEFEF}\textit{\textcolor{blue}{-0.5}} & \cellcolor[HTML]{EFEFEF}\textit{\textcolor{blue}{-0.6}} & \cellcolor[HTML]{EFEFEF}\textit{\textcolor{blue}{-0.5}} & \cellcolor[HTML]{EFEFEF}\textit{\textcolor{blue}{-0.3}} & \cellcolor[HTML]{EFEFEF}\textit{\textcolor{blue}{-0.7}} & \cellcolor[HTML]{EFEFEF}\textit{\textcolor{blue}{-0.4}} & \cellcolor[HTML]{EFEFEF}\textit{\textcolor{blue}{-0.3}} \\ \hline \noalign{\hrule height 0.5pt}
\end{tabular}
* Improvement is computed against the best baseline (excluding ours).
\end{table}

In Table~\ref{tab:quadsyn_exp_pck-add}, \textsf{QuadPose} achieves the best PCK@$2.5$/5/10 and AUC, as well as the lowest median error for 2D keypoint localization.
The gap to DREAM~\cite{lee2020camera} is particularly large at PCK@$2.5$ (24.45 vs. 92.18), indicating that methods without skeleton-aware modeling are highly sensitive to self-occlusion and leg appearance ambiguity.
SGTAPose~\cite{tian2023robot} and YOLOv8-pose~\cite{yolov8-pose} perform well under relaxed PCK thresholds (5/10\,px) but degrade at the stricter 2.5\,px threshold, whereas \textsf{QuadPose} maintains high precision by combining keypoint-specific features with skeleton-aware kinematic constraints.

\textsf{QuadPose} achieves the highest ADD accuracy across all thresholds, demonstrating robust 3D base pose estimation.
DREAM~\cite{lee2020camera} remains weak at strict thresholds (e.g., 19.1\% at ADD@10cm), SGTAPose~\cite{tian2023robot} benefits from temporal context (52.08\%) but is still affected by limb ambiguity, RoboPose~\cite{labbe2021single} struggles with complex articulations, and RoboPEPP~\cite{goswami2025robopepp} degrades under stricter criteria.
Overall, \textsf{QuadPose} maintains stable 3D pose estimation by coupling precise 2D keypoint localization with temporal modeling.

\begin{figure}[t]
    \centering
    \includegraphics[width=1\linewidth]{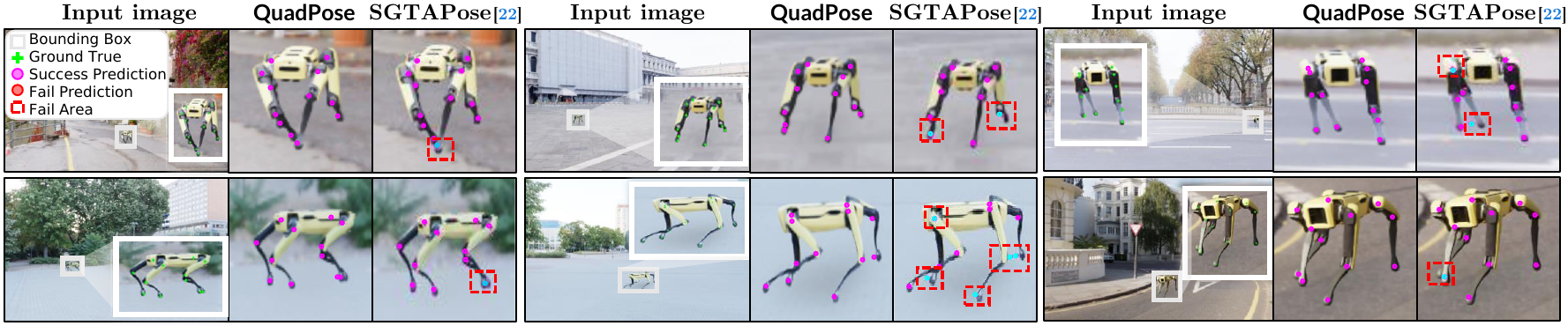}
    \caption{\textbf{Qualitative comparison of 2D keypoint detection on \texttt{QuadSyn}.}}
    \label{fig:07}
\end{figure}

\begin{figure}[t]
    \centering
    \includegraphics[width=1\linewidth]{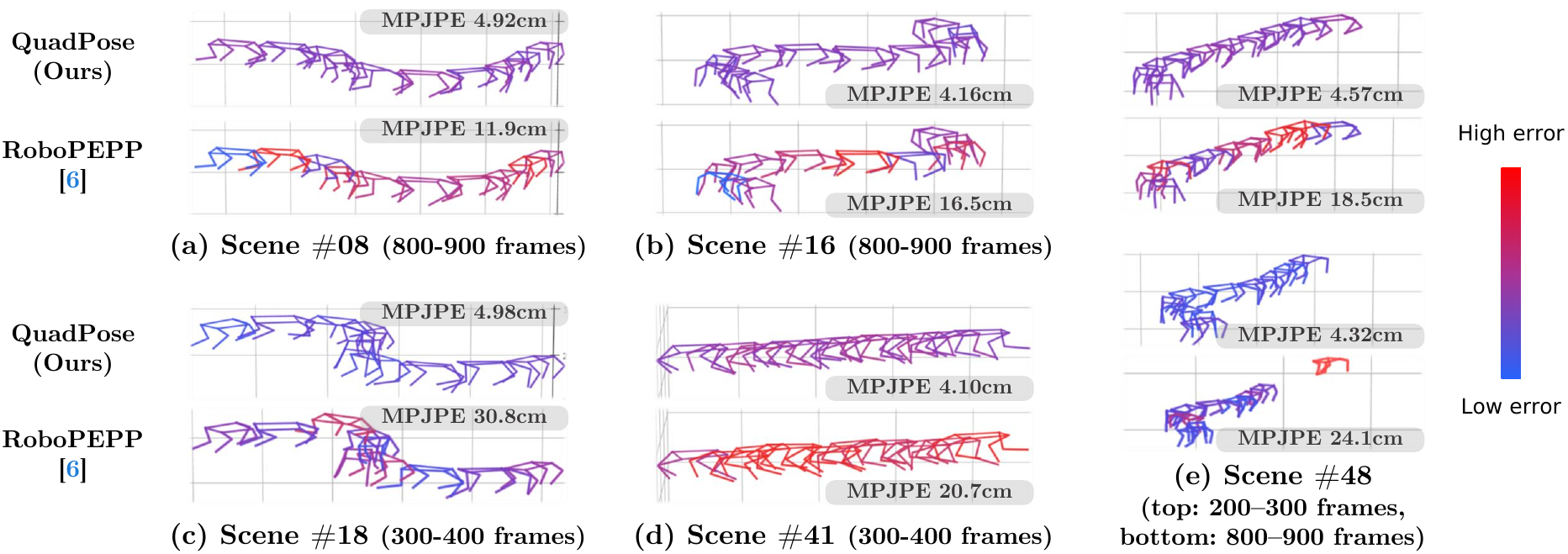}
    \caption{\textls[-20]{\textbf{Visualization of 3D pose estimation results on \texttt{QuadSyn}.}
        \textsf{QuadPose} (ours) vs. RoboPEPP~\cite{goswami2025robopepp}. Colors indicate error. MPJPE (Mean Per Joint Position Error) in cm.}}
    \label{fig:08}
\end{figure}

\textls[-20]{
Table~\ref{tab:quadsyn_exp_joint} reports the joint-angle error for all 12 joints; \textsf{QuadPose} achieves the lowest error on 11 out of 12 joints, with a mean error of 3.65$^\circ$.
RoboPose~\cite{labbe2021single} exhibits errors above 8$^\circ$ across all joints, indicating that mesh-alignment methods struggle to handle articulated motion.
RoboPEPP~\cite{goswami2025robopepp} shows a slight advantage at J3 (front-left foot), which may be related to larger motion at the end-effector joint.
Overall, \textsf{QuadPose} delivers the best performance, indicating that skeleton-aware masked self-attention helps mitigate error propagation along the kinematic chain.
}

Figure~\ref{fig:07} compares 2D keypoint predictions of \textsf{QuadPose} and the strongest baseline, SGTAPose~\cite{tian2023robot}, on \texttt{QuadSyn}.
\textsf{QuadPose} localizes keypoints more precisely, especially under self-occlusion and symmetric leg configurations where SGTAPose tends to confuse leg identities.
Figure~\ref{fig:08} compares 3D pose estimation results of \textsf{QuadPose} and RoboPEPP~\cite{goswami2025robopepp}, showing that \textsf{QuadPose} consistently achieves lower MPJPE across diverse viewpoints and gait patterns.

\subsection{Ablation Study}
\label{sec:exp:ablation}

\begin{table}[t]
\centering
\begin{minipage}[t]{0.48\textwidth}
    \centering
    \renewcommand{\arraystretch}{1.0}
    \scriptsize
    \setlength{\tabcolsep}{2pt}
    \caption{\textls[-10]{\textbf{Ablation study} on the 2D keypoint detector with different configurations.}}
    \label{tab:ablation:keypoint}
    \begin{tabular}{lccccc}
    \hline \noalign{\hrule height 0.5pt}
     & \multicolumn{5}{c}{\textbf{PCK @ (pix)}} \\
    \multirow{-2}{*}{\textbf{Settings}} & \textbf{2.5}\up & \textbf{5}\up & \textbf{10}\up & \textbf{AUC}\up & \textbf{Med}\down \\ \hline \noalign{\hrule height 0.5pt}
    Encoder & 80.99 & 95.03 & 99.29 & 91.37 & 1.200 \\
    + SimCC & 88.98 & 96.82 & 99.35 & 93.78 & 0.736 \\
    + ST & 91.77 & \textbf{97.83} & \textbf{99.68} & 95.11 & 0.540 \\
    \rowcolor[HTML]{EFEFEF}
    + SA & \textbf{92.18} & 97.57 & 99.58 & \textbf{95.54} & \textbf{0.407} \\ \hline \noalign{\hrule height 0.5pt}
    \end{tabular}
\end{minipage}
\hfill
\begin{minipage}[t]{0.48\textwidth}
    \centering
    \renewcommand{\arraystretch}{1.0}
    \scriptsize
    \setlength{\tabcolsep}{1pt}
    \caption{\textbf{Ablation study} on the 3D pose estimator with different configurations.}
    \label{tab:ablation:pose}
    \begin{tabular}{lccccc}
    \hline \noalign{\hrule height 0.5pt}
     & \multicolumn{5}{c}{\textbf{ADD @ (cm)}} \\
    \multirow{-2}{*}{\textbf{Settings}} & \textbf{10}\up & \textbf{15}\up & \textbf{20}\up & \textbf{AUC}\up & \textbf{Med}\down \\ \hline \noalign{\hrule height 0.5pt}
    Base & 45.40 & 69.23 & 79.87 & 54.78 & 10.76 \\
    + Joint Frozen & 53.80 & 74.55 & 82.35 & 59.33 & 9.360 \\
    + Temporal & 67.47 & \textbf{81.12} & \textbf{85.94} & 65.90 & 7.230 \\
    \rowcolor[HTML]{EFEFEF}
    + Refinement & \textbf{69.49} & \textbf{81.12} & \textbf{85.94} & \textbf{66.98} & \textbf{6.740} \\ \hline \noalign{\hrule height 0.5pt}
    \end{tabular}
\end{minipage}
\end{table}

To analyze the contribution of each component in the keypoint detector, we conduct an ablation study.
Table~\ref{tab:ablation:keypoint} summarizes the results for four configurations:
\begin{itemize}
    \renewcommand{\labelitemi}{$\bullet$}
    \item \textbf{Encoder}: A baseline model that predicts 2D keypoints directly from encoder features using a simple MLP head.
    \item \textbf{+ SimCC}: Add a SimCC head that predicts independent 1D distributions for the $x$ and $y$ axes, replacing direct coordinate regression.
    \item \textbf{+ ST (selector token)}: Introduce learnable selector tokens that attend to encoder tokens to extract keypoint-specific representations.
    \item \textbf{+ SA (skeleton-aware masked attention)}: Apply skeleton-aware masked self-attention that constrains attention to physically connected joints.
\end{itemize}

SimCC yields the largest improvement over the encoder-only baseline (PCK@$2.5$: $+$7.99, median error: $-$0.46 px), enabling more robust localization than direct regression.
The selector tokens (+ST) further improve accuracy (PCK@$2.5$: 91.77, AUC: 95.11), while skeleton-aware masked self-attention (+SA) improves temporal stability (AUC: 95.54, median error: 0.41 px).
The slight drop in PCK@5/10 (97.83$\rightarrow$97.57, 99.68$\rightarrow$99.58) occurs near saturation, as masked attention shifts borderline predictions toward skeleton-consistent locations.

Table~\ref{tab:ablation:pose} ablates the 3D pose estimator across four configurations:
\begin{itemize}
    \renewcommand{\labelitemi}{$\bullet$}
    \item \textbf{Base}: Jointly trains the JointNet and Base Pose Head from a single frame, without temporal modeling or refinement.
    \item \textbf{+ Joint Frozen}: Freezes JointNet and trains only the Base Pose Head.
    \item \textbf{+ Temporal}: Adds 1D temporal convolutions over $T$ consecutive frames to leverage inter-frame motion cues.
    \item \textbf{+ Refinement}: Adds an MLP-based refinement module that predicts residual offsets to correct accumulated upstream errors.
\end{itemize}

Freezing the joint predictor (+Joint Frozen) is the most critical design choice (ADD@10: 45.40$\rightarrow$53.80, AUC: 54.78$\rightarrow$59.33), confirming that joint training causes error coupling that destabilizes the FK chain; the supplementary material shows that freezing reduces the mean joint error from 11.55$^\circ$ to 3.65$^\circ$.
Temporal modeling (+Temporal) yields the largest gain (ADD@10: 53.80$\rightarrow$67.47, AUC: 59.33$\rightarrow$65.90) by resolving monocular depth ambiguity through multi-frame context.
The refinement head (+Refinement) adds gains at strict thresholds (ADD@10: 67.47$\rightarrow$69.49, AUC: 65.90$\rightarrow$66.98) and lowers the median error to 6.74\,cm.

\subsection{Results on Real-world Data: \texttt{Harper}}
\label{sec:harper}

\begin{figure}[t]
\centering
\begin{minipage}[t]{0.38\linewidth}
    \vspace{0pt}
    \centering
    \captionof{table}{\textbf{Overall accuracy and temporal stability on real-world \texttt{Harper}~\cite{avogaro2024harper} test set.}
    We report OKS mAP and temporal jitter ($\uparrow$: higher is better, $\downarrow$: lower is better).}
    \label{tab:real-world}
    {\renewcommand{\arraystretch}{0.9}
    \resizebox{\linewidth}{!}{%
    \begin{tabular}{lcc}
    \hline \noalign{\hrule height 0.5pt}
    \textbf{Method} & \textbf{\begin{tabular}[c]{@{}c@{}}OKS\\ mAP\up\\ (Avg.)\end{tabular}} & \textbf{\begin{tabular}[c]{@{}c@{}}Temporal\\ Jitter\down\\ (Avg.)\end{tabular}} \\ \hline \noalign{\hrule height 0.5pt}
    DREAM~\cite{lee2020camera} & 14.79 & \textbf{12.53} \\
    YOLOv8~\cite{yolov8-pose} & \textul{60.77} & 171.5 \\
    SGTAPose~\cite{tian2023robot} & 0.000 & 334.1 \\
    \rowcolor[HTML]{EFEFEF}
    \textbf{\textsf{QuadPose} (Ours)} & \textbf{64.13} & \textul{24.32} \\
    \rowcolor[HTML]{EFEFEF}
    \textit{Improvement*} & \textit{\textcolor{red}{+3.36}} & \textit{\textcolor{red}{+11.8}} \\ \hline \noalign{\hrule height 0.5pt}
    \end{tabular}}}
    {\scriptsize *Improvement is relative to the best baseline (excluding ours).}
\end{minipage}
\hfill
\begin{minipage}[t]{0.57\linewidth}
    \vspace{10pt}
    \centering
    \includegraphics[width=\linewidth]{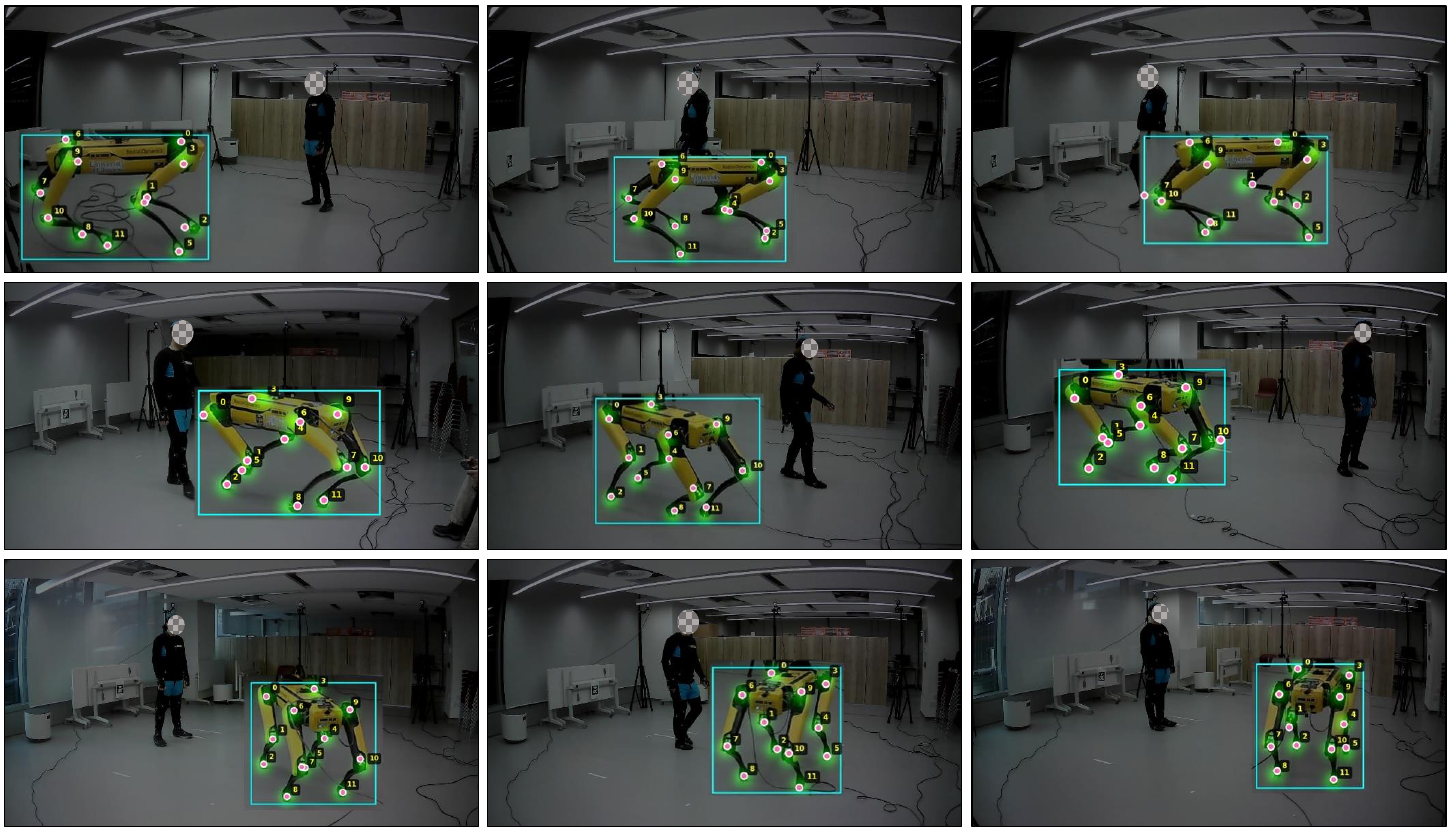}
    \caption{\textls[-30]{\textbf{Real-world visualization of 2D keypoint detection on \texttt{Harper}.}
    Zoomed-in robot views with \textcolor{green}{$\bullet$} manual 2D references and \textcolor{pink}{$\bullet$} \textsf{QuadPose} predictions.}}
    \label{fig:real}
\end{minipage}
\end{figure}

To evaluate sim-to-real transfer, we test on \texttt{Harper}~\cite{avogaro2024harper}, a real-world dataset captured with a fixed external camera.
Unlike \texttt{QuadSyn}, \texttt{Harper} provides neither camera intrinsics nor robot 3D annotations, so we limit evaluation to 2D keypoint accuracy and temporal stability.
We therefore manually annotate 2D keypoints as references and visualize them as Gaussian heatmaps to account for pixel-level labeling ambiguity.

\textls[-15]{
Table~\ref{tab:real-world} reports OKS mAP and temporal jitter on the real-world test set.
\textsf{QuadPose} outperforms YOLOv8 by +3.36 in OKS mAP (64.13) and significantly reduces temporal jitter from 171.5 to 24.32, showing temporally stable predictions.
Note that low jitter alone does not imply accuracy: DREAM~\cite{lee2020camera} has the lowest jitter (12.53) yet poor mAP (14.79), indicating inaccurate predictions, while SGTAPose~\cite{tian2023robot} exhibits near-zero mAP with high jitter, suggesting overfitting to synthetic appearances.
\textsf{QuadPose} is the only method that achieves both high accuracy and low jitter.
}

Figure~\ref{fig:real} presents qualitative results.
Despite training solely on synthetic data without explicit domain adaptation, predicted keypoints consistently fall within or near the manual reference regions, confirming that the learned representation generalizes across the sim-to-real domain gap.
Combined with the known robot kinematics, accurate 2D keypoints directly enable reliable 3D pose recovery.
Failure case analysis is provided in the supplementary material.

\section{Conclusions}
\label{sec:conclusion}

We presented \textsf{QuadPose}, a camera-to-quadruped full-body pose estimation framework that recovers the 18-DoF state from external monocular RGB video sequences, and introduced \texttt{QuadSyn}, the first benchmark dataset for this task with full-body annotations.
By improving 2D keypoint localization under leg symmetry and self-occlusion and leveraging temporal modeling, \textsf{QuadPose} achieves the best performance on \texttt{QuadSyn}.
Our keypoint-focused design (selector tokens and skeleton-aware masked attention) improves robustness under challenging test scenarios, enabling reliable 3D pose recovery.
Qualitative results on \texttt{Harper} suggest promising sim-to-real transfer in external-camera monitoring scenarios.
Future work includes extending \texttt{QuadSyn} to more diverse motions and robot platforms (e.g., Go1, ANYmal), and collecting real-world 3D ground truth.

\bibliographystyle{splncs04}
\bibliography{main}

\end{document}